\documentclass{article} 
\usepackage{iclr2023_conference,times}

\usepackage{amsmath,amsfonts,bm}

\def\eqref#1{equation~\ref{#1}}

\def\1{\bm{1}}

\DeclareMathAlphabet{\mathsfit}{\encodingdefault}{\sfdefault}{m}{sl}
\SetMathAlphabet{\mathsfit}{bold}{\encodingdefault}{\sfdefault}{bx}{n}

\usepackage{hyperref}
\usepackage{url}
\usepackage{bm}             
\usepackage[pdftex]{graphicx} 
\usepackage{float}          
\usepackage{multirow}       
\usepackage{booktabs}       
\usepackage{amsmath}        
\usepackage{makecell}       
\usepackage{caption} 

\title{Signal Is Harder To Learn Than Bias: \\ Debiasing with Focal Loss}

\author{
  Moritz Vandenhirtz, Laura Manduchi, Ri\v{c}ards Marcinkevi\v{c}s and Julia E. Vogt \\
  \ Department of Computer Science, ETH Zurich\\
  \small{\texttt{\{moritz.vandenhirtz, laura.manduchi, ricardsm, julia.vogt\}@inf.ethz.ch
}}}

\iclrfinalcopy 

\begin{document}

\maketitle

\begin{abstract} 
Spurious correlations are everywhere. While humans often do not perceive them, neural networks are notorious for learning unwanted associations, also known as biases, instead of the underlying decision rule. As a result, practitioners are often unaware of the biased decision-making of their classifiers. Such a biased model based on spurious correlations might not generalize to unobserved data, leading to unintended, adverse consequences.
We propose Signal is Harder (SiH), a variational-autoencoder-based method that simultaneously trains a biased and unbiased classifier using a novel, disentangling reweighting scheme inspired by the focal loss. Using the unbiased classifier, SiH matches or improves upon the performance of state-of-the-art debiasing methods. To improve the interpretability of our technique, we propose a perturbation scheme in the latent space for visualizing the bias that helps practitioners become aware of the sources of spurious correlations.

\end{abstract}

\section{Introduction} 
The generalization capability of deep neural networks (DNN) highly depends on the quality of the training data. If spurious correlations are present, the model might ignore the intrinsic \textit{signal attributes} while still performing reasonably well in classification tasks. However, such a biased model will not be robust and will not generalize outside the training distribution. 
To increase the trustworthiness of machine learning algorithms and prevent unwanted consequences, it is crucial to avoid deploying biased models~\citep{geirhos2020shortcut}. Thus, there has been an increased interest in the community to mitigate this problem. While many methods assume and utilize an observed variable that captures the source of bias for each data point, recently, some effort has been made to alleviate this prohibitive assumption~\citep{nam2020learning}. 

For example, consider a dataset comprising of images of vehicles. A DNN might implicitly use the \textit{bias attribute} ``sky'' as a shortcut for classifying planes because most images of airplanes are shot while they are in the air. Throughout the paper, we will call samples \textit{bias-aligned} when their bias attributes are strongly correlated with the label. Here, leveraging the bias as a decision rule leads to the correct predicted label, e.g. airplane in the sky. Conversely, \textit{bias-conflicting} data points are the samples for which the biased decision rule leads to the wrong prediction, e.g. aircraft in the hangar.

Recent efforts by \citet{nam2020learning} aim to eliminate the need for an observed variable that captures the source of bias for each data point. They assume that malignant bias attributes are easier to learn than the underlying signal. Based on this easy-to-learn assumption, they train a biased classifier that focuses on the easy, bias-aligned samples. Simultaneously, they train an unbiased classifier by upweighting the remaining hard, bias-conflicting samples. We propose an alternative reweighting for the unbiased classifier based on the focal loss~\citep{focal_loss} that does not require the previously utilized subtle, distorting stability measures. We motivate the usage of this loss function through the easy-to-learn assumption, which infers that signal is harder to learn than bias. 

In addition, we extend the literature by integrating a variational autoencoder~(VAE)~\citep{Kingma} into the model. At inference time, this allows us to make use of latent perturbations to remove the biasing attributes from the embeddings, which we then feed to the decoder to visualize debiased images. Comparing these images with the original reconstructions can help practitioners uncover unknown biases.

\paragraph{Contribution}
We propose a novel reweighting scheme, coined Signal is Harder~(SiH), for training an unbiased classifier.\footnote{Our code is publicly available at \url{https://github.com/mvandenhi/Signal-is-Harder}} Due to the lack of labels for the unknown bias, SiH exploits the assumption that signal is harder to learn than bias and utilizes a reweighting based on the well-established focal loss~\citep{focal_loss}. We show that this direct mechanism improves the debiasing capabilities compared to the existing, more complex reweighting scheme by \citet{nam2020learning}. Additionally, by training a VAE simultaneously with the classifiers, the unknown bias can be visualized in the reconstructions. For this, the proposed algorithm perturbs the latent bias embeddings at inference time to remove the bias without creating artifacts in the reconstructions. We improve upon previous methods as our minimal perturbation does not change other aspects of the reconstruction, unambiguously 
unveiling 
the unknown spurious attribute.

\section{Related work}

\paragraph{Separating samples by difficulty}
Recent works separate bias-conflicting from bias-aligned samples to train an unbiased classifier~\citep{nam2020learning,lee2021learning,kim2021biaswap}. This separation can be achieved by differentiating data points through the difficulty of predicting their label. In a standard classification setting, \citet{zhang2018generalized} propose the Generalized Cross Entropy (GCE) loss to reduce the weight on samples whose labels are hard to predict:
\begin{equation}
     GCE(\hat{y},y) = \frac{1-\hat{y}^q}{q},
    \label{GCE_original}
\end{equation}
where $\hat{y}$ is the predicted probability of the correct label $y$ according to the classifier, and $q\in(0,1]$ is a hyperparameter to control the strength of emphasis. 
The GCE is best understood by inspecting its derivative
    $\frac{\partial GCE(\hat{y},y)}{\partial \boldsymbol{\theta}} = \hat{y}^q \frac{\partial CE(\hat{y},y)}{\partial \boldsymbol{\theta}}$,
where $\boldsymbol{\theta}$ are the learnable neural network parameters. 
This loss upweighs samples that the classifier already predicts well, ignoring samples for which the current decision rule does not work. Contrary to the GCE loss, the Focal Loss (FL) by \citet{focal_loss} puts more focus on hard, misclassified examples:
\begin{equation}
     FL(\hat{y},y) = (1-\hat{y})^q CE(\hat{y},y)
    \label{FL}
\end{equation}
With this reweighting scheme, the samples whose labels are hard to predict are upweighted such that the classifier does not ignore the samples for which finding a decision rule is a hard problem.

\paragraph{Debiasing without supervision}
Previous works focused on predictions with respect to known sensitive attributes~\citep{sagawa2019distributionally,edwards2015censoring,Learningnottolearn}, which are often difficult to retrieve. For this reason \citet{nam2020learning} propose LfF, a new approach to debias a classifier, which does not require bias attributes. They assume that bias is only malignant if it is easier to learn than the true signal attribute and leverage the GCE loss to focus on the easy, bias-aligned samples to train a biased classifier. Simultaneously, they train an unbiased classifier, designed to learn the true, underlying signal. For this they upweigh the bias-conflicting samples, i.e. the data points for which the bias can not be utilized to predict the label, by the relative difficulty score (RDS) 
\begin{equation}
    RDS(\hat{y}_s,\hat{y}_b,y) = \frac{CE(\hat{y}_b,y)}{CE(\hat{y}_s,y)+CE(\hat{y}_b,y)},
    \label{RDS}
\end{equation}
where $\hat{y}_s$ and $\hat{y}_b$ are the predicted probabilities of the correct label $y$ according to the unbiased and biased classifier, respectively. However, before inserting the CE terms into the above formula, they apply an empirically motivated exponential moving average and a class-wise normalization by the maximum CE to each term. In the following section, we will propose an enhanced upweighting mechanism for the unbiased classifier that does not require weight-distorting stability measures. 

\citet{lee2021learning} extend the method of \citet{nam2020learning} by additionally swapping the learned latent bias embeddings of different inputs to decouple the bias from the label. At inference time, they train a decoder to visualize the embeddings with and without swapped bias, such that the unknown bias can be discovered by analyzing the differences between the two reconstructions. To avoid misleading artifacts in the visualization, we will propose a more conservative perturbation, which relies on a VAE trained simultaneously with the classifiers.
Further discussion can be found in Appendix~\ref{further related work}.

\section{Method} 
\begin{figure}[htbp]
    \centering
    \includegraphics[width=0.6\linewidth]{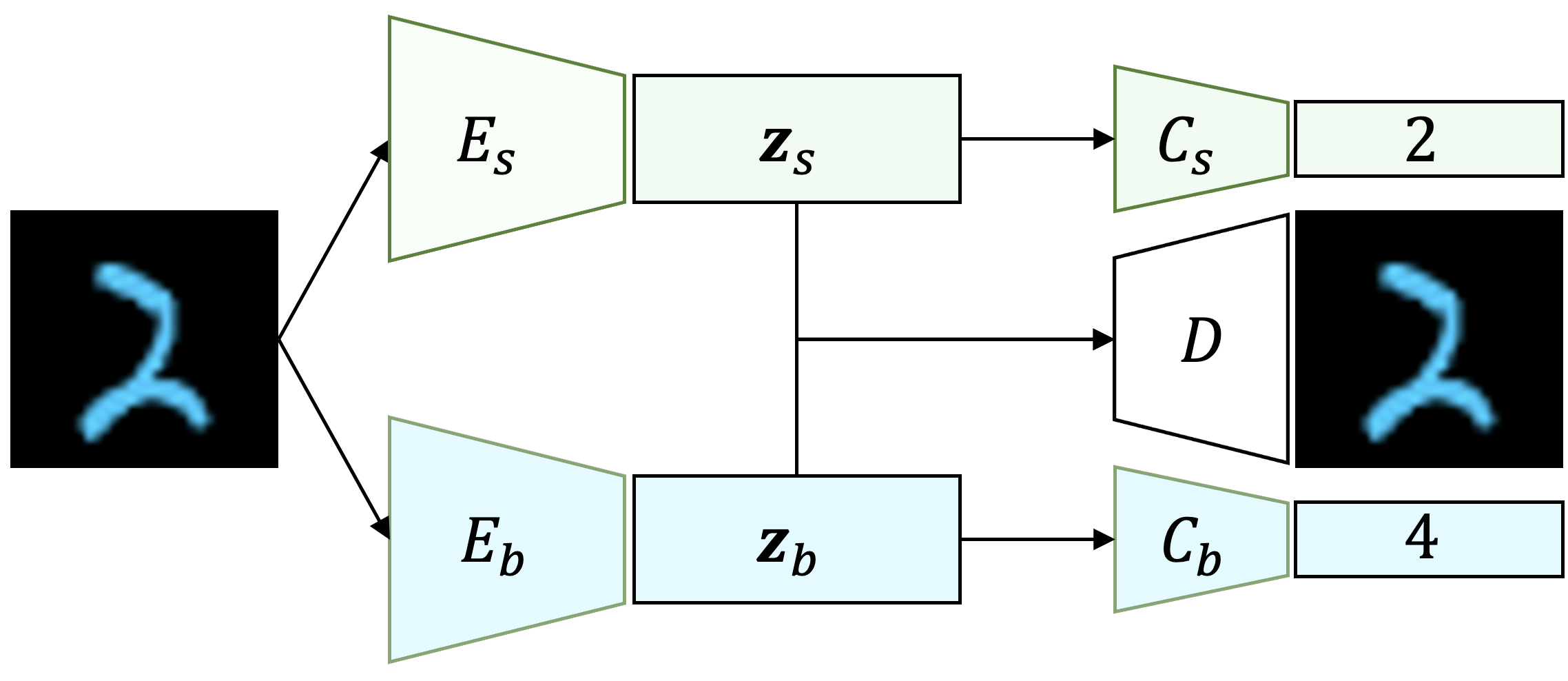}
    \caption{Graphical overview of our model's structure for a bias-conflicting image. The input $\textbf{x}$ is passed through the signal and bias encoders $E_s $ and $ E_b$ to obtain the latent signal and bias embeddings $\textbf{z}_s$ and $\textbf{z}_b$, which in this example should be the digit \textit{two} and the color \textit{blue}, respectively. These representations are then passed through their respective classifier $C_s$ and $C_b$ to predict the label. Lastly, $\textbf{z}_s$ and $\textbf{z}_b$ are concatenated and passed through the decoder $D$ to reconstruct the image.}
    \label{Our Model}
\end{figure}

We propose a debiasing algorithm, coined Signal is Harder (SiH), consisting of a VAE-based architecture and a new weighting mechanism for training the unbiased classifier. The VAE uses two encoders to map the input into signal and bias embeddings, which are concatenated and passed through the decoder to reconstruct the original input. Additionally, we train an unbiased and biased classifier on signal and bias embeddings, respectively. The biased classifier is trained by upweighting bias-aligned samples through the GCE loss. In contrast, the unbiased classifier is trained by upweighting bias-conflicting samples through our novel focal-loss-based weighting scheme, which we will introduce in the next paragraph. The generative nature of the model allows us to produce bias visualizations that help discover the unknown source of bias. We depict the proposed model structure in Figure~\ref{Our Model}.


\paragraph{Reweighting by focal loss}
We hereby introduce a new reweighting scheme, aiming to utilize the easy-to-learn assumption not only for the biased but also for the unbiased classifier. Similar to the GCE that upweighs bias-aligned samples for the biased classifier, we want to utilize a mirrored loss function that upweighs the remaining bias-conflicting samples for the unbiased classifier. 


The aforementioned reasons motivate the inclusion of the focal loss for training the unbiased classifier. We use this loss to identify samples for which the biased classifier struggles to predict the correct class and emphasize these presumably bias-conflicting samples by upweighting them when training the unbiased classifier. As the information learned by the biased classifier should leverage the unbiased classifier, but not the other way around, we detach the weighting factor from the computational graph during backpropagation and obtain the following update for the unbiased classifier:
\begin{equation}
    \frac{\partial \mathcal{L}_s(\hat{y}_s,\hat{y}_b,y)}{\partial \boldsymbol{\theta}_s} = (1-\hat{y}_b)^q \frac{\partial CE(\hat{y}_s,y)}{\partial \boldsymbol{\theta}_s},
\end{equation}
where $q\in(0,1]$ is a hyperparameter controlling the strength of emphasis.
With this loss, we exploit that bias-conflicting samples are hard to learn for a biased classifier. By focusing on these data points, the unbiased classifier is forced to learn the signal, as here, leveraging the bias does not lead to the correct prediction.
Most importantly, the straightforward integration of the focal loss for training the unbiased classifier removes the need for weight-distorting stability measures.

\paragraph{Latent adversarial perturbation}
To make practitioners aware of the unknown spurious correlations in a dataset, we propose a visualization approach by perturbing the bias at inference time. We perturb the bias embeddings, such that the bias contained within is removed, and reconstruct the debiased image. We argue that such a perturbation should be as small as possible, such that no artifacts are created in the process since a practitioner needs to consider every change to the input as a potential bias. To achieve this, we adapt and utilize the adversarial perturbations from Deepfool~\citep{moosavi2016deepfool}. This algorithm is designed to find a minimal perturbation to the input that fools the classifier into predicting the wrong class. Thus, we perturb the bias representations such that the biased classifier can no longer predict the correct class; effectively, the perturbation removes the bias from the bias embeddings. 

Having trained a VAE, we can use the decoder at inference time to visualize the perturbed bias embeddings together with the unchanged signal representation. By comparing the original reconstruction with the debiased visualization, it is possible to identify the spurious correlations in the image.
In contrast to DisEnt, we train the decoder simultaneously with the classifiers to encode all image-relevant information in the latent representations, thus, supporting the unbiased classifier in finding the signal attributes and improving reconstruction quality.

\section{Experiments} 
To compare the performance of our method, SiH, with previous works, we evaluate it on Colored MNIST~\citep{Learningnottolearn} and Corrupted CIFAR-10~\citep{cifar10c} with a varying percentage of bias-conflicting images during the training. For a detailed description of the datasets, we refer to Appendix~\ref{dataset visualizations}. We determine three baselines to which we compare the proposed approach. The first baseline we implement is a Vanilla model consisting of one encoder and classifier, which measures the standard performance without any debiasing scheme.
The second model we compare the proposed approach to, is LfF from \citet{nam2020learning}. 
Lastly, we compare SiH to DisEnt by \citet{lee2021learning}, a recently proposed state-of-the-art debiasing algorithm, which also visualizes the bias.

\subsection{Quantitative evaluation}
\label{quantitative evaluation}
\paragraph{Comparison on test sets}
In Table~\ref{table results}, we show the performance of all models on the unbiased test set of Colored MNIST and Corrupted CIFAR-10. The estimates differ from the values presented in the baseline papers~\citep{nam2020learning,lee2021learning} because we also vary random seeds over dataset generation instead of only over the weight initialization. 

\begin{table}[t!]
\caption{Unbiased test set accuracy + standard deviation in \%. The method with the significantly highest accuracy is denoted in \textbf{bold}. Otherwise, insignificantly different methods are \underline{underlined}.}
\vspace{1mm}
\centering
\scriptsize
\vspace{-0.25cm}
\begin{tabular}{crcccc}\toprule 
Dataset & Ratio & Vanilla & LfF & DisEnt & SiH\\
\midrule
 \multirow{6}{*}{Colored MNIST} & 20\% \qquad & \textbf{94.92} {\scriptsize $\pm$ 0.24} & 70.18 {\scriptsize $\pm$ 4.19}  & 90.94 {\scriptsize $\pm$ 1.46} & 85.24 {\scriptsize $\pm$ 1.60}  \\ 

 & 10\% \qquad & \textbf{91.24} {\scriptsize $\pm$ 0.26} & 81.99 {\scriptsize $\pm$ 5.01}  & 89.12 {\scriptsize $\pm$ 1.44} & 85.35 {\scriptsize $\pm$ 1.23}  \\ 

 & 5\% \qquad  & \underline{85.48} {\scriptsize $\pm$ 0.50} & 	81.18 {\scriptsize $\pm$ 2.94}  & \underline{85.54} {\scriptsize $\pm$ 2.49} & \underline{86.14} {\scriptsize $\pm$ 1.78}  \\ 

 &2\% \qquad  & 73.28 {\scriptsize $\pm$ 0.56} & 76.97 {\scriptsize $\pm$ 2.49} & \underline{82.38} {\scriptsize $\pm$ 1.68} & \underline{83.80} {\scriptsize $\pm$ 1.28}  \\

 &1\% \qquad  & 59.41 {\scriptsize $\pm$ 0.39} & 68.91 {\scriptsize $\pm$ 5.01} & 76.33 {\scriptsize $\pm$ 3.41} & \textbf{80.03} {\scriptsize $\pm$ 2.04} \\

 &0.5\% \qquad  & 43.70 {\scriptsize $\pm$ 0.83} & 60.42 {\scriptsize $\pm$ 2.72} & 63.98 {\scriptsize $\pm$ 4.78} & \textbf{71.63} {\scriptsize $\pm$ 2.49} \\
 \midrule
 \multirow{6}{*}{Corrupted CIFAR-10}&20\% \qquad & \underline{67.57} {\scriptsize $\pm$ 0.41} & 64.50 {\scriptsize $\pm$ 2.17}  & 60.99 {\scriptsize $\pm$ 5.84} & \underline{66.75} {\scriptsize $\pm$ 1.34} \\ 

 &10\% \qquad & 57.11 {\scriptsize $\pm$ 0.76} & \underline{59.29} {\scriptsize $\pm$ 3.16}  & 53.47 {\scriptsize $\pm$ 4.43} & \underline{61.26} {\scriptsize $\pm$ 2.06}  \\ 

 &5\% \qquad & 46.89 {\scriptsize $\pm$ 0.78} & \underline{55.77} {\scriptsize $\pm$ 2.33}  & 46.40 {\scriptsize  $\pm$ 5.81} & \underline{55.63} {\scriptsize  $\pm$ 1.54} \\ 

 &2\% \qquad & 34.90 {\scriptsize $\pm$ 0.81} & \underline{47.26} {\scriptsize $\pm$ 1.56} & 36.98 {\scriptsize  $\pm$ 4.43} & \underline{43.66} {\scriptsize  $\pm$ 1.81}  \\

 &1\% \qquad & 28.22 {\scriptsize $\pm$ 0.73} & \textbf{39.39} {\scriptsize $\pm$ 2.16} & 31.22 {\scriptsize  $\pm$ 2.69} & 35.17 {\scriptsize  $\pm$ 1.19} \\

 &0.5\% \qquad & 22.26 {\scriptsize $\pm$ 1.03} & \underline{30.04} {\scriptsize $\pm$ 1.67} & \underline{31.97} {\scriptsize  $\pm$ 3.34} & 27.30 {\scriptsize  $\pm$ 2.04} \\
\bottomrule
\end{tabular}
\label{table results}
\vspace{-0.25cm}
\end{table}

We observe that Vanilla outperforms the debiasing algorithms for the 10\% and 20\% cases of Colored MNIST. Thus, the easy-to-learn assumption is likely not fulfilled for these training sets. The debiasing methods show their benefit only for a lower amount of bias-conflicting samples. Here, SiH outperforms or at least matches all baselines, while DisEnt is the runner-up. Especially for the 0.5\% setting, there is a considerable gap in performance between our and other methods.

For Corrupted CIFAR-10, the best-performing models are LfF and SiH. We observe that for higher percentages of bias-conflicting samples, SiH is better than LfF, while for lower proportions, the opposite is the case. DisEnt seems to be generally worse than the other debiasing methods. Finally, Vanilla performs worse than the debiasing methods except for the 20\% case. Thus, the debiasing methods present an improvement over a standard empirical risk minimizer.

\paragraph{Focal loss vs. RDS weighting}

\begin{figure}[t]
\vspace{-0.25cm}
\begin{minipage}{\textwidth}
\begin{minipage}[b]{0.55\textwidth}
\centering
\includegraphics[width=1\linewidth]{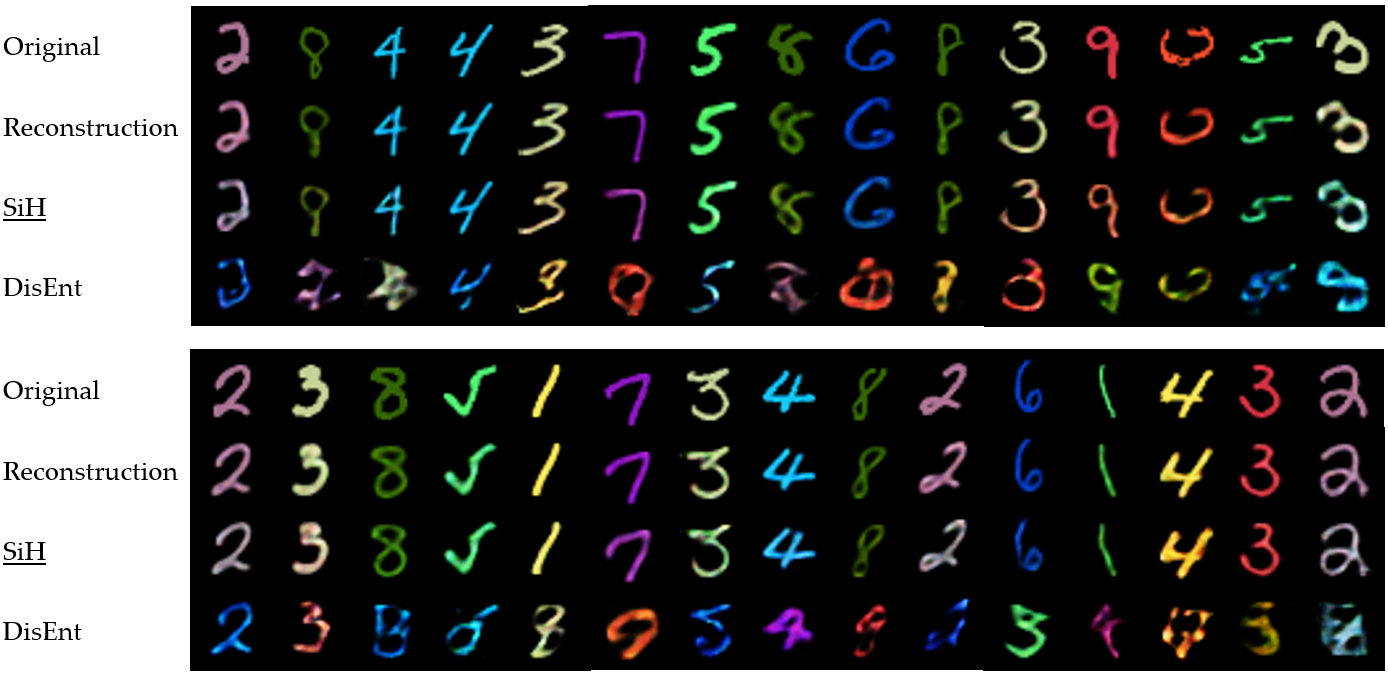}
\captionof{figure}{A random collection of bias visualizations for Colored MNIST. The randomly selected images are varied over random seeds and the percentage of bias-conflicting images in the training set.}
\label{cmnist bias}
\end{minipage}
\hfill
\begin{minipage}[b]{0.42\textwidth}
\captionof{table}{Unbiased accuracy + standard deviation in \% for Colored MNIST and Corrupted CIFAR-10. 
}
\centering
\scriptsize
\begin{tabular}{crcc}\toprule 
Dataset & Ratio & SiH$_{RDS}$ & SiH\\
\midrule
 \multirow{6}{*}{\makecell[l]{Colored\\MNIST}} & 20\% \qquad & 80.15 {\scriptsize $\pm$ 5.27} & \textbf{85.24} {\scriptsize $\pm$ 1.60}  \\ 

 & 10\% \qquad &  \underline{86.13} {\scriptsize $\pm$ 3.07} & \underline{85.35} {\scriptsize $\pm$ 1.23}  \\ 

 & 5\% \qquad & \underline{84.10} {\scriptsize $\pm$ 3.03} & \underline{86.14} {\scriptsize $\pm$ 1.78}  \\ 

 &2\% \qquad  & 79.38 {\scriptsize $\pm$ 2.37} & \textbf{83.80} {\scriptsize $\pm$ 1.28}  \\

 &1\% \qquad  & 74.22 {\scriptsize $\pm$ 3.21} & \textbf{80.03} {\scriptsize $\pm$ 2.04} \\

 &0.5\% \qquad & 64.17 {\scriptsize $\pm$ 5.74} & \textbf{71.63} {\scriptsize $\pm$ 2.49} \\
 \midrule
 \multirow{6}{*}{\makecell[l]{Corrupted\\CIFAR-10}}&20\% \qquad & \underline{64.09} {\scriptsize $\pm$ 5.64} & \underline{66.75} {\scriptsize $\pm$ 1.34} \\ 

 &10\% \qquad & 56.89 {\scriptsize $\pm$ 5.09} & \textbf{61.26} {\scriptsize $\pm$ 2.06}  \\ 

 &5\% \qquad & 51.25 {\scriptsize  $\pm$ 4.10} & \textbf{55.63} {\scriptsize  $\pm$ 1.54} \\ 

 &2\% \qquad & 38.22 {\scriptsize  $\pm$ 3.77} & \textbf{43.66} {\scriptsize  $\pm$ 1.81}  \\

 &1\% \qquad & 31.64 {\scriptsize  $\pm$ 2.50} & \textbf{35.17} {\scriptsize  $\pm$ 1.19} \\

 &0.5\% \qquad & 24.59 {\scriptsize  $\pm$ 1.84} & \textbf{27.30} {\scriptsize  $\pm$ 2.04} \\
\bottomrule
\end{tabular}
\label{table ablation}
\end{minipage}
\end{minipage}
\vspace{-0.25cm}
\end{figure}

In Table~\ref{table ablation}, we show the performance on the unbiased test set of Colored MNIST and Corrupted CIFAR-10 using two different ways of weighting the samples for training the unbiased classifier. While SiH stands for our proposed method, in SiH$_{RDS}$, we utilize the RDS proposed by \citet{nam2020learning} when reweighting the data points. 
The results suggest that on average our proposed reweighting mechanism significantly increases performance while the inclusion of the VAE leads to an accuracy-interpretability tradeoff. Additionally, the focal loss reduces the variability in accuracy across multiple runs. 

Overall, the quantitative results show that the proposed reweighting scheme improves performance. For settings where the easy-to-learn assumption is likely to be fulfilled, SiH shows promising results compared to baselines. The ablation study shows that integrating our reweighting for the unbiased classifier is critical in improving its accuracy. 

\subsection{Qualitative evaluation}
\label{qualitative evaluation}
Figure~\ref{cmnist bias} displays the bias visualization from DisEnt and SiH for a few randomly selected images. Additionally, in Figure~\ref{appendix random cifar} of Appendix~\ref{app cifar10c bias}, we show the random bias visualizations for Corrupted CIFAR-10. We will not analyze the latter images further, as here, signal and bias are not disentangled well enough for visualizing the bias for either method.

For Colored MNIST, the swapping of DisEnt perturbs the bias representations so strongly that this also leads to an unwanted change in the digit. This change is likely due to the bias and signal representations not being perfectly disentangled. Thus, the leftover signal in the bias dimensions gets swapped too. On the other hand, SiH does not perturb the digit while regularly perturbing the color. However, due to the weaker magnitude of change, our approach sometimes does not visibly change the image.  

SiH is more conservative when generating perturbations, which is advantageous for visualizing bias in realistic cases where learned signal and bias embeddings are not perfectly disentangled. Although our changes are more subtle, we believe that, for a practitioner, our perturbation method should be preferred, as it does not induce artifacts, which otherwise have to be considered as a possible bias. 

\section{Conclusion and future work}
In the presence of bias, a classifier often leverages these spurious correlations rather than the underlying signal. The application of such an algorithm can have adverse consequences in critical situations.
This work advances the research in building unbiased deep learning models by investigating a novel reweighting scheme. We propose SiH, which trains a bias classifier to be as biased as possible and simultaneously trains an unbiased classifier by upweighting samples for which the biased decision rule fails to predict the correct labels. We show that the proposed weighting factor based on the focal loss can match or outperform existing works. Additionally, by training a generative model, users are able to visualize and identify the bias at inference time. For this, the proposed approach leverages latent adversarial perturbations that do not introduce undesirable artifacts. 

\paragraph{Future work}
Although SiH has demonstrated its effectiveness on simple datasets, its efficacy on more challenging datasets and other modalities requires further investigation. For this, it is crucial to use more expressive generative models such as generative adversarial networks~\citep{generative_adv_networks} or diffusion models~\citep{diffusion1,diffusion2,diffusion3}. Moreover, while SiH consists of established individual components, their combination is not rigorously derived. In fact, the entire field would profit from greater mathematical rigor, beginning with the establishment of a theoretical definition of what constitutes the ``ease of learning''.

\newpage


\bibliography{iclr2023_conference}
\bibliographystyle{iclr2023_conference}

\newpage
\appendix

\section{Further related work}
\label{further related work}

\paragraph{Debiasing without explicit supervision}
To train an unbiased classifier, several existing approaches~\citep{bahng2020learning,clark2019don,wang2019learning,RUBi} leverage some form of implicit knowledge about the bias present in the dataset. They use this insight to design and train a model architecture susceptible to the specific bias attribute. They then train an unbiased model to learn decision rules different from the biased model. For example, \citet{bahng2020learning} capture texture bias in image classification by training a convolutional neural network with a small receptive field. Simultaneously, they train an unbiased model by forcing it to learn representations that are independent of the biased ones.

\paragraph{Debiasing without implicit supervision}
\citet{liu2021just} slightly adapt the idea of \citet{nam2020learning}. Instead of training both classifiers simultaneously, they divide the training into two stages. First, the biased classifier is trained. Second, they train the unbiased classifier and upweigh all samples that were misclassified by the biased classifier. Hence, they also try to focus on training on the bias-conflicting samples. 

DisEnt by \citet{lee2021learning} builds upon \citet{nam2020learning} by utilizing their training scheme to create a disentangled representation useful for feature augmentation. As motivation, they show that the diversity of training samples is an important factor in training. Instead of training the unbiased classifier only on the sparse bias-conflicting samples, they try to synthesize additional samples for which using the bias as decision rule does not work. Their algorithm works as follows: First, \citet{lee2021learning} train the base structure from \citet{nam2020learning} using GCE and RDS to create disentangled representations where signal and bias dimensions can be separated. After a predetermined number of updates, they start to swap the bias dimensions of different samples to synthesize representations with the same signal but different bias. Hence, making the bias unusable as decision rule for the label because it originates from a different sample. They then train their classifiers on those representations as well as on the original samples. 

A caveat of \citet{nam2020learning} is the absence of a mechanism for visualizing the unknown bias.
For this, \citet{lee2021learning} propose to train a decoder ex-post to reconstruct the images given their latent representations. By repeating their swapping process and reconstructing the images after the swap, they expect to visualize the bias present in the dataset as they can compare reconstructions with the same signal but swapped bias dimensions. This will work if the learned embeddings are perfectly disentangled but might introduce artifact if they are not.

Further research for debiasing and visualizing the bias has been done by \citet{darlow2020latent}, which leverage a vector quantized variational autoencoder~\citep{van2017neural} and add a simple biased classifier on top of the latent representations. After training, they perturb the latent dimensions such that the biased classifier is as unsure as possible about the label. This perturbed representation is then passed through the decoder to generate images without relevant bias. Consequently, a second, unbiased classifier is trained on these images, which should not contain bias that can be used for predicting the label.

\section{Implementation details}

In line with previous works~\citep{nam2020learning,lee2021learning,kim2021biaswap}, we utilize an MLP for Colored CMNIST for all methods. Each encoder consists of three linear layers with a bottleneck of size 100 for signal and bias, respectively. The decoder is again consisting of three linear layers. As activation function, we use the Rectified Linear Unit (ReLU). For the classifiers, we solely use one linear layer for both datasets. This is because for us, the difficulty of a dataset is determined by the complexity of the connection between latent variables and the realization $\textbf{x}$ thereof. If we knew the latent variables, inferring the label would be simple. Thus, a single linear layer suffices. 

We adapt the encoder-decoder structure for Corrupted CIFAR-10 from an MLP to a CNN, where we use a ResNet18~\citep{Resnet} with bottleneck dimension of 512 for the encoders of all methods and a ResNet18-like decoder. 

We do not perform any preprocessing for Colored MNIST. For the preprocessing of Corrupted CIFAR-10, we take random crops consisting of at least $50\%$ of the original image and resize them to the original $32\times32$ size. Additionally, we allow horizontal flips of the images and standardize the pixel values over the entire dataset. To calculate the reconstruction loss, we transform the standardized pixel values back into $[0,1]$ so that its size is comparable among all datasets.

To have visualizations that capture the original image well, for SiH, we upweigh the reconstruction loss by the factor 100. Additionally, we rescale the reconstruction and KL term by dividing through $3L$, where $L$ is the number of pixels, to be invariant to image resolution and number of channels while retaining their relative loss magnitude. 

For updating the model weights, we use the Adam optimizer~\citep{Adam} with a learning rate of 0.001 and batch size of 256 for both datasets. The hyperparameter $q$ is chosen to be $0.7$ by following the GCE coiners \citet{zhang2018generalized}.

We perform early stopping and reduce the learning rate when plateauing by computing this loss function on a held-out 10\% of the training set. For Colored MNIST, we use an early stop patience of 2 versus 20 for Corrupted CIFAR-10. The patience for the learning rate reduction is one-half of the early stop patience and reduces the learning rate by a factor of 10. 

We do not perform hyperparameter tuning on specific settings, as there must not be an unbiased or bias labelled validation set in the setting of unknown bias. We would like to encourage future work to do the same.

For creating visualization of the bias, we adapt Deepfool~\citep{moosavi2016deepfool} for our purposes. Originally, the algorithm was developed for perturbing pixels in an input image, while we use it for perturbing latent dimensions. Thus, while pixel values need to be clamped in $[0,1]$, we do not require this. For the distance measure of the perturbation, we use the $\ell_2-$norm.

For the bias visualization, we use the trained model of SiH as backbone. This is vital in ensuring that discrepancies in the visualizations of DisEnt and SiH can be attributed solely to the differing visualization techniques. We perturb images for which the biased and unbiased classifiers predict the correct class. With this, we aim to find bias-aligned images, which we can then perturb into neutral images. We randomly sample a different target class and apply our perturbation. For DisEnt, we pick a second image, for which the biased classifier predicts the same sampled target class and swap the bias embeddings.

\section{Dataset visualizations}
\label{dataset visualizations}

To compare the performance of our method SiH with previous works, we evaluate it on two datasets. These datasets consist of a training set for which we define varying percentages of bias-conflicting images to analyse the performance. We assess the debiasing potential of all methods on an unbiased test set where signal and bias are independently and uniformly distributed. For SiH, we do not perform hyperparameter tuning on each framework because an unbiased validation set does not exist in the setting of unknown bias.

\begin{figure}[H]
    \centering
    \includegraphics[width=\linewidth]{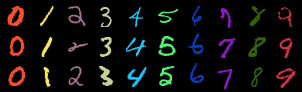}
    \caption{Bias-aligned images of the Colored MNIST dataset. The columns show different digits $\textbf{z}_s$ with their respective colors $\textbf{z}_b$ that predominantly manifest in combination.}
    \label{CMNIST_fig}
\end{figure}

\begin{figure}[H]
    \centering
    \includegraphics[width=\linewidth]{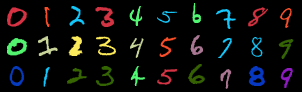}
    \caption{Bias-conflicting images of the Colored MNIST dataset. The columns show different digits $\textbf{z}_s$ with colors $\textbf{z}_b$ that are usually not observed together.}
    \label{CMNIST_fig_confl}
\end{figure}

\begin{figure}[H]
    \centering
    \includegraphics[width=\linewidth]{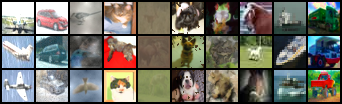}
    \caption{Bias-aligned images of the Corrupted CIFAR-10 dataset. The columns show the different classes $\textbf{z}_s$ with their respective corruptions $\textbf{z}_b$ that predominantly manifest in combination.
    For example, the class birds often has foggy images, while ships are frequently pixelated.}
    \label{CIFAR_fig}
\end{figure}

\begin{figure}[H]
    \centering
    \includegraphics[width=\linewidth]{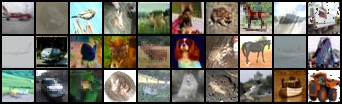}
    \caption{Bias-conflicting images of the Corrupted CIFAR-10 dataset. The columns show the different classes $\textbf{z}_s$ with corruptions $\textbf{z}_b$ that are usually not observed together.}
    \label{CIFAR_fig_confl}
\end{figure}

The first dataset is Colored MNIST by \citet{Learningnottolearn}, which consists of the popular handwritten digit database MNIST~\citep{Lecun_mnist} synthetically infused with a color bias. To each digit, we randomly assign distinct mean colors, which serve as bias attributes. Hence, the signal $\textbf{z}_s$ is the digit while the easy-to-learn bias $\textbf{z}_b$ manifests itself as the color. In Figure~\ref{CMNIST_fig}, we display bias-aligned images, for which leveraging the color as decision rule would lead to the correct label. A minority of samples in the training set consists of bias-conflicting samples, showed in Figure~\ref{CMNIST_fig_confl}, for which the biased decision rule leads to the wrong prediction. Learning to recognize the digit instead of the color is the only valid decision rule with which bias-aligned as well as bias-conflicting samples can be correctly classified.
The second dataset we apply SiH to is the Corrupted CIFAR-10 dataset~\citep{cifar10c}. It is based on the standard CIFAR-10 dataset~\citep{krizhevsky2009learning}, injected with synthetically generated corruptions such as fog, brightness, or saturation for each class. These synthetic biasing perturbations are designed to be as realistic as possible. A collection of randomly selected bias-aligned and bias-conflicting images for both datasets can be found in Figure~\ref{CIFAR_fig} and Figure~\ref{CIFAR_fig_confl}, respectively.

\newpage


\newpage

\section{Bias visualizations for Corrupted CIFAR-10}
\label{app cifar10c bias}
\begin{figure}[H]
    \centering
    \includegraphics[width=\linewidth]{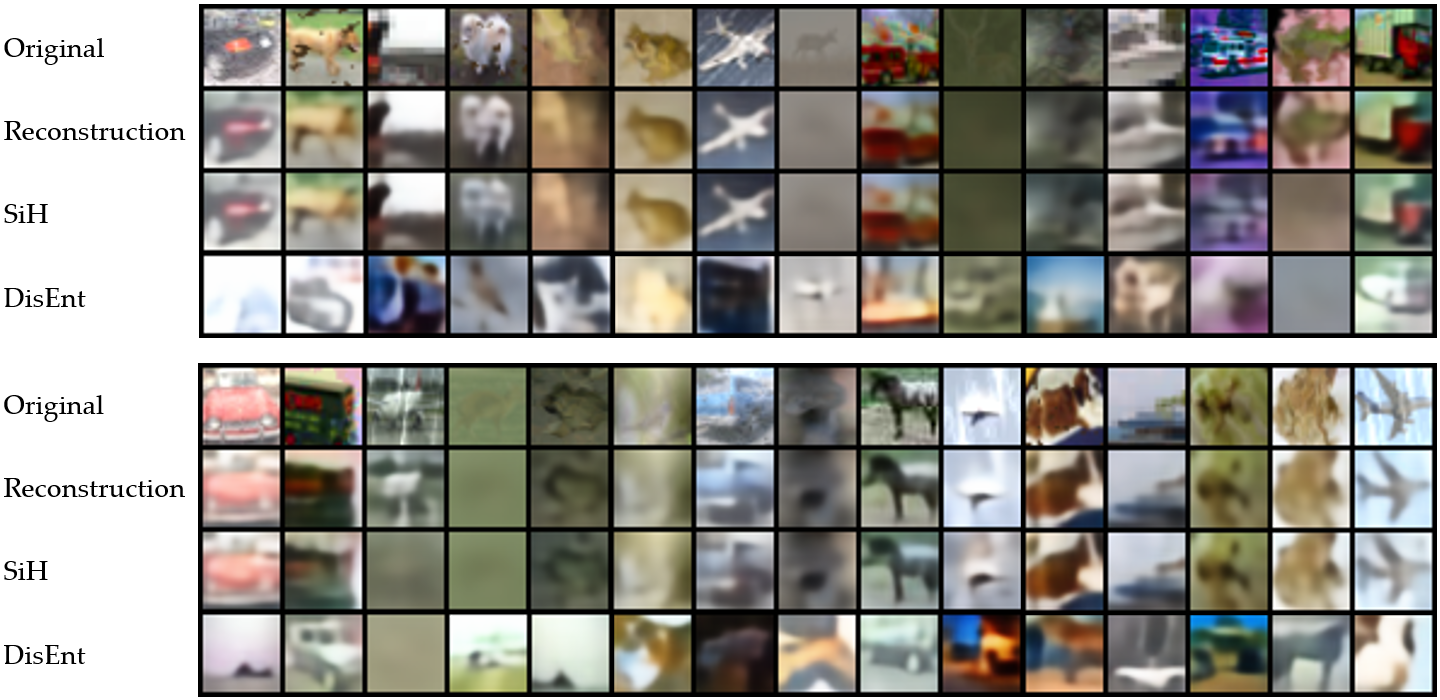}
    \caption{A random collection of bias visualizations for Corrupted CIFAR-10. The randomly selected images are varied over random seeds and the percentage of bias-conflicting images in the training set.}
    \label{appendix random cifar}
\end{figure}

\end{document}